\documentclass{article}

\usepackage[preprint]{neurips_2026}

\usepackage{xcolor}         
\definecolor{linkColor}{rgb}{0.2,0.4,0.6}
\usepackage[utf8]{inputenc} 
\usepackage[T1]{fontenc}    
\usepackage[colorlinks=true,linkcolor=linkColor,citecolor=linkColor,filecolor=linkColor,urlcolor=linkColor]{hyperref}       
\usepackage{url}
\usepackage{booktabs}
\usepackage{amsfonts}
\usepackage{amsmath}
\usepackage{nicefrac}
\usepackage{microtype}
\usepackage{graphicx}
\usepackage{multirow}
\usepackage{xspace}
\usepackage{enumitem}
\usepackage{CJKutf8}
\usepackage{fancyvrb}

\newcommand{\name}{\texorpdfstring{\textsc{OfficeEval}}{OfficeEval}\xspace}

\title{Mind the Gap: Can Frontier LLMs Pass a Standardized Office Proficiency Exam?}

\author{%
  Microsoft Research\\
  \href{https://aka.ms/GeneralAI}{https://aka.ms/GeneralAI}
}

\begin{document}

\maketitle

\begin{abstract}
The deployment of Large Language Model (LLM) agents for computer automation is accelerating, yet their ability to navigate complex, professional-grade productivity software is largely untested. We argue that Office automation is an ideal environment for benchmarking document-automation capability, as it requires long-horizon planning and reasoning, precise parameter configuration, and multi-application integration. To quantify this capability, we introduce an evaluation based on China's National Computer Rank Examination (NCRE), featuring 200 comprehensive practical-operation tasks across Word, Excel, and PowerPoint. Each task is scored on a 100-point rubric scale using 7,118 machine-gradable criteria, and Score Rate (SR) denotes the mean percentage of rubric points earned across these tasks. We benchmark 7 frontier LLMs and observe stark limitations: single-turn models score a maximum of 36.6\%. A stronger agentic system with execution feedback, iterative repair, and broader Office automation access reaches 68.8\%, but remains below the 95.5\% community-reference score used as a scoring sanity check. Ultimately, our experiments demonstrate that despite recent advancements in code generation, achieving reliable fine-grained Office document automation remains a significant challenge for current code-generating LLM and agent systems.
\end{abstract}

\section{Introduction}
\label{sec:intro}



The transition toward autonomous Large Language Model (LLM) agents has driven rapid progress in domains ranging from software engineering to web navigation \citep{jimenez2024swebench,zhou2024webarena,xie2024osworld,mialon2024gaia}. However, if these agents are to serve as viable ``digital workers'', they must master the environments where human work actually takes place. Office suites remain among the most widely used environments for knowledge work, and proficiency across applications like Word, Excel, and PowerPoint remains a foundational workplace skill. Yet, despite its real-world importance, agent evaluation in Office automation remains surprisingly underdeveloped. Current studies rely heavily on synthetic environments, narrow single-application slices, or subjective LLM-as-a-judge scoring, none of which adequately capture the complexity of genuine Office workflows.

To establish a rigorous standard for Office-based agents, we propose evaluating them against the exact same practical examinations used to certify human professionals. In this paper, we introduce a novel benchmarking framework based on China's National Computer Rank Examination (NCRE), a massive-scale standardized testing system that has evaluated over 110 million candidates \citep{neea2024ncrefaq}. Leveraging an established human certification resolves several major benchmarking bottlenecks simultaneously. The NCRE provides a suite of foundational and advanced tasks that are designed by domain experts, carefully calibrated for difficulty, and crucially evaluated using objective, machine-readable scoring rubrics. Furthermore, unlike traditional QA or isolated code-completion tasks, a single NCRE problem requires an agent to execute dozens of distinct, interdependent operations on real-world documents. This transforms a human professional exam into an ideal, highly challenging testbed for evaluating long-horizon sequential decision-making.

\begin{figure}[t]
  \centering
  \includegraphics[width=\linewidth]{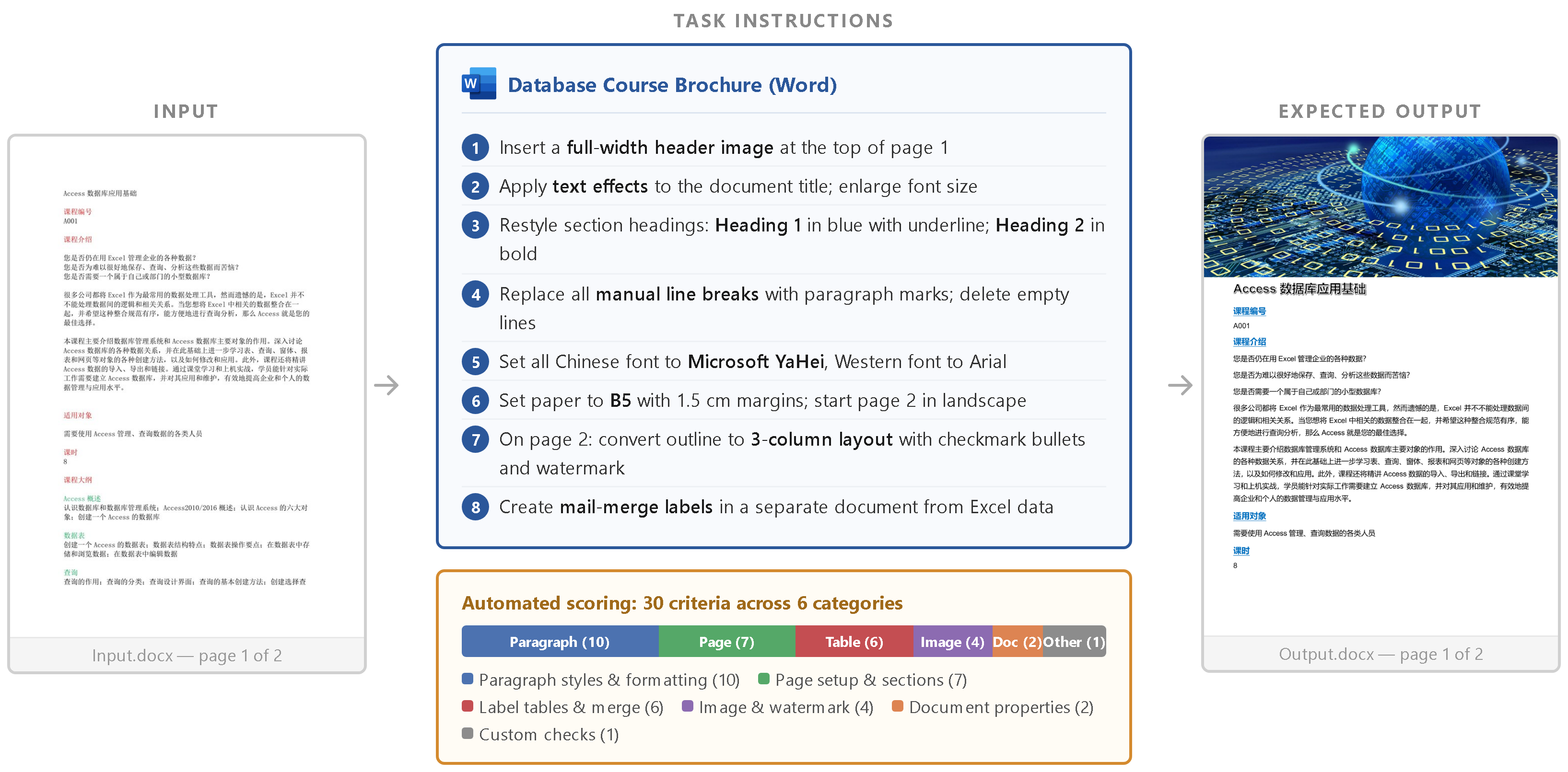}
  \caption{End-to-end illustration of a Word task in \name. The original document (\textbf{left}) is transformed according to the task instructions (\textbf{center}) into a styled brochure with header image, heading styles, and mail-merge labels (\textbf{right}). Only page~1 of the 2-page document is shown; several steps (e.g., 3-column layout, watermark) apply to page~2. The task is scored by 30 deterministic criteria across 6 skill categories. Instructions are translated from the original Chinese; additional examples across Word, Excel, and PowerPoint appear in the Appendix.}
  \label{fig:tasks}
\end{figure}

Specifically, we introduce \name, a comprehensive benchmark constructed from 200 NCRE tasks and evaluated against 7,118 machine-gradable criteria across Word, Excel, and PowerPoint. Using this benchmark, we systematically evaluate 7 frontier LLMs under two paradigms: single-turn code generation and autonomous coding-agent systems.
Our primary contributions and key findings are summarized as follows:
\begin{itemize}[nosep, leftmargin=*]
\item \textbf{A benchmark grounded in real-world professional certification.} By transforming the NCRE into \name, we provide an evaluation framework anchored by a 60-point reference threshold for the extracted practical subset and a 95.5\% community-reference score used as a scoring sanity check. This establishes a highly realistic and quantifiable baseline for long-horizon Office automation under a standardized practical-task rubric.
\item \textbf{Systematic evaluation revealing a critical performance gap.} Our evaluation demonstrates that no single-turn model reaches the 60-point reference threshold on the extracted practical subset, with the best model (Claude Opus 4.7) achieving only a 36.6\% Score Rate (SR). While autonomous coding agents show substantial improvements where Claude Code reaches 53.0\% and Codex reaches 68.8\%, all models remain below the community-reference score. Furthermore, performance is highly polarized (dropping to a 2.8\% SR for Grok-4.1-fast), indicating that frontier API status alone does not guarantee strong Office automation performance.
\item \textbf{Fine-grained diagnosis and error taxonomy.} To understand \emph{why} models fail, we introduce a criterion-level taxonomy that isolates execution crashes from logic errors. This reveals a key insight: in the stronger coding-agent setting, code execution success rises from under 50\% to 98--99\%, but operation accuracy remains low. Current agents can successfully write code that runs, but still struggle to implement the correct Office-specific semantic operations.
\end{itemize}

\section{Related Work}
\label{sec:related}

LLM agent benchmarks now span web navigation~\citep{zhou2024webarena,deng2024mind2web}, software engineering~\citep{jimenez2024swebench}, multi-environment reasoning~\citep{liu2024agentbench,mialon2024gaia}, and desktop automation~\citep{xie2024osworld,bonatti2024windowsarena,xu2025theagentcompany}. To our knowledge, no prior benchmark combines real standardized Office-exam tasks with deterministic criterion-level grading across Word, Excel, and PowerPoint.

Within Office automation, existing work covers individual applications or narrow scopes.
Word tasks appear as minor components in broader suites~\citep{xie2024osworld,wang2024officebench,mu2025gui360} without dedicated formatting evaluation.
Spreadsheet benchmarks~\citep{li2024sheetcopilot,ma2024spreadsheetbench,chen2024sheetagent} focus on formula and data manipulation, under-representing chart customization, pivot tables, and conditional formatting.
Presentation benchmarks~\citep{guo2023pptc,huang2025pptbench} cover layout and editing but omit animations, transitions, and cross-application skills.
The most comparable multi-application efforts, OfficeBench~\citep{wang2024officebench} and OdysseyBench~\citep{wang2025odysseybench}, study Office workflows across applications, but their evaluation targets workflow-level task completion rather than NCRE-style deterministic, criterion-level grading of fine-grained document properties.

\name differs from prior work in two respects: \textbf{(1)} tasks come from NCRE, a nationally administered certification exam designed by domain-expert committees, providing externally validated difficulty and broad skill coverage that synthetic or crowdsourced tasks cannot match; and \textbf{(2)} all 7,118 scoring criteria are machine-gradable, enabling deterministic, fine-grained evaluation across Word, Excel, and PowerPoint without LLM or human judging variance. A further consequence of \textbf{(1)} is that scores on \name carry an externally defined meaning. The per-criterion point allocations are taken directly from the NCRE task rubrics, so a model's SR is reported on the same per-task 100-point rubric scale used for the extracted practical-operation tasks. It is the share of allocated points the model earned on this subset, giving the score an external rubric anchor rather than only a benchmark-internal ranking among systems.

\section{\name}
\label{sec:benchmark}

\subsection{Data source}
\label{sec:data_source}

\name is derived from the practical operation component of China's National Computer Rank Examination (NCRE), specifically the Level~1 and Level~2 MS Office modules~\citep{neea2025ncrelevel1,neea2025ncrelevel2}.
NCRE is a nationally standardized proficiency test administered by China's National Education Examinations Authority and designed to assess practical computing skills~\citep{neea2024ncrefaq}.
Level~1 evaluates foundational Office skills such as basic formatting, simple formulas, and standard presentation creation, while Level~2 covers more advanced operations including mail merge, pivot tables, chart customization, and complex animations~\citep{neea2025ncrelevel1,neea2025ncrelevel2}.
The full NCRE exam also includes multiple-choice questions on computer fundamentals and, at Level~1, basic OS and internet tasks; we extract only the \emph{practical Office operation} sections (Word, Excel, and PowerPoint), which account for the majority of the exam score (60\% at Level~1, 80\% at Level~2) and constitute the primary assessment of hands-on Office proficiency~\citep{neea2025ncrelevel1,neea2025ncrelevel2}.
This two-level structure provides a natural difficulty gradient for analyzing how model performance varies with task complexity.

The NCRE certificate is a nationally recognized credential. The full NCRE exam combines multiple-choice questions on computer fundamentals with the practical-operation tasks studied in this work, and a candidate passes (and receives a certificate from the Ministry of Education) by scoring at least 60 out of 100 \emph{on the full exam}~\citep{neea2024ncreresults}. Our \name benchmark extracts the practical-operation subset. Every scoring criterion and its point allocation is taken from the task rubric, so SR is the share of allocated points the model earned and is reported on the same per-task 100-point rubric scale for these practical tasks, a property that benchmarks built from synthetic or crowdsourced tasks cannot offer. Throughout this paper, the 60-point value serves as a reference threshold for subset-level score interpretation.

Task instructions are in Chinese, which is well-supported by all multilingual LLMs evaluated in this work.
To investigate whether instruction language affects model performance, we also constructed English-translated versions of all 200 tasks---including document content, scoring criteria, and font/style mappings---for a cross-language analysis (Section~\ref{sec:discussion}).
We provide English translations of example tasks in the appendix for reference.\footnote{The appendix examples are drawn from official NCRE sample materials publicly posted by the National Education Examinations Authority: \url{https://ncre.neea.edu.cn/xhtml1/category/1507/848-1.htm}.}

\paragraph{Scope and data availability.}
The NCRE practical examination tasks, input documents, and scoring configurations are copyrighted examination materials authored by the National Education Examinations Authority and exam-preparation publishers. We therefore do not redistribute the raw task statements, original input documents, or scoring scripts. To make the study reproducible, we provide in the appendix the full set of prompts we used, the experimental environment and settings, the evaluation procedure, and the criterion-level statistics, so that the same pipeline can be reproduced on independently obtained NCRE materials or applied to similar Office-proficiency data sources.

\subsection{Task format}
\label{sec:task_format}

Each task in \name consists of three components:
\begin{enumerate}[nosep, leftmargin=*]
    \item \textbf{Input document:} The initial document file (Word \texttt{.docx}, Excel \texttt{.xlsx}, or PowerPoint \texttt{.pptx}), along with any supporting materials such as images, data files, or theme templates that may be referenced in the instructions.
    \item \textbf{Task instructions:} Natural language descriptions of the required operations, typically 5--15 distinct sub-tasks of varying complexity. Instructions often include \emph{reference images} that illustrate the target formatting, layout, or style (see Figure~\ref{fig:tasks} for representative examples). This is a natural consequence of Office work: complex formatting goals (table borders, chart styles, SmartArt layouts) are inherently visual and often easier to convey through an image than through text alone. This also mirrors real workplace practice; a coworker or designer may hand over only a style mock-up and expect the document to be produced to match. The ability to translate a visual style into concrete Office operations is itself a practical skill being assessed. The benchmark is therefore inherently \textbf{multimodal}: solving a task often requires interpreting visual references alongside textual descriptions.
    \item \textbf{Scoring configuration:} A machine-readable XML file defining all evaluation criteria, including the properties to check, expected values, comparison operators, and point allocations.
\end{enumerate}

The evaluation pipeline is independent of how the output document is produced: given the task instructions and input document, any method produces an output document, which is then scored automatically against the scoring configuration.

\begin{figure}[t]
  \centering
  \includegraphics[width=\linewidth]{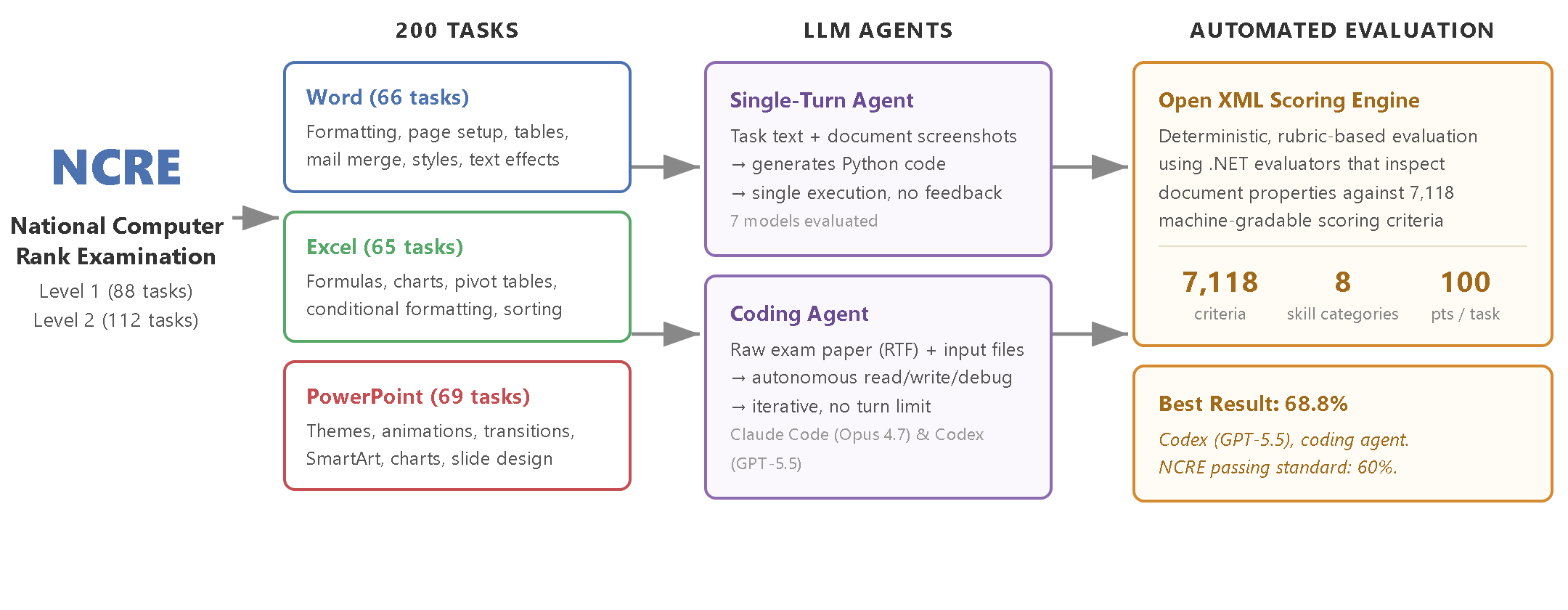}
  \caption{Overview of the \name benchmark and evaluation pipeline. 200 exam tasks from NCRE span three applications and two difficulty levels. Tasks are processed by either a single-turn code generator or an autonomous coding agent, and output documents are scored by a deterministic evaluation engine against 7,118 fine-grained criteria.}
  \label{fig:overview}
\end{figure}

Figure~\ref{fig:overview} illustrates the overall task format and evaluation pipeline.

\subsection{Evaluation framework}
\label{sec:eval}

Our evaluation engine parses Office Open XML documents standardized by ISO/IEC 29500~\citep{iso29500} using the Open XML SDK~\citep{microsoft2024openxmlsdk} and runs as a set of Windows executables (.NET, x86).
Most scoring criteria are evaluated by parsing the Open XML document structure directly; however, a subset of checks (e.g., certain chart renderings and complex formatting validations) invoke Microsoft Office COM automation internally, so a Windows environment with Microsoft Office installed is required for full evaluation.
The scoring logic is specified by configuration files and deterministic evaluator code. For a fixed compatible Windows, Office, font, and locale environment, the same input document yields the same score.

Each task's scoring configuration defines a set of \emph{scoring criteria}, each specifying a criterion type (e.g., \texttt{PageSetup}, \texttt{Chart}, \texttt{PivotTable}), one or more evaluation rules with expected values and comparison operators, and a point value (on a 100-point scale).
Figure~\ref{fig:xml_example} shows a representative criterion from one task's configuration: it checks that the font of the text ``\begin{CJK*}{UTF8}{gbsn}指标体系构建\end{CJK*}'' is set to ``\begin{CJK*}{UTF8}{gbsn}华文新魏\end{CJK*}'' and awards 0.8 points if so. A complete task typically contains 20--70 such criteria covering every operation in the rubric.

\begin{figure}[t]
\centering
\small
\begin{CJK*}{UTF8}{gbsn}
\begin{Verbatim}[commandchars=\\\{\}]
<ScorePointDefine WepName="Paragraph find/replace"
                  WepType="ParaFind">
  <MarkRule FullScore="0.8">
    <FindParams PropKey="Font" PropKeyDesc="Font"
                Find="指标体系构建"/>
    <StdAnswer><Value>华文新魏</Value>
      <ValueDesc>Title font incorrect;
        should be 华文新魏.</ValueDesc>
    </StdAnswer>
    <OperatorParam ValueType="0">0</OperatorParam>
  </MarkRule>
</ScorePointDefine>
\end{Verbatim}
\end{CJK*}
\caption{Representative scoring criterion from a Word task. The criterion locates the text ``\begin{CJK*}{UTF8}{gbsn}指标体系构建\end{CJK*}'', checks that its font equals ``\begin{CJK*}{UTF8}{gbsn}华文新魏\end{CJK*}'', and awards 0.8 points on a match. Each task contains 20--70 such criteria.}
\label{fig:xml_example}
\end{figure}

Each task is scored on a 100-point scale, with the total distributed across its scoring criteria in fine-grained increments.
This rubric-based approach enables \emph{partial credit scoring}: an agent that correctly completes 70\% of the required operations receives roughly 70\% of the points, providing a much more informative signal than binary success/failure metrics.

The evaluation engine is implemented as a command-line tool: users submit an output document and task ID, and receive a structured JSON result containing the total score and per-criterion pass/fail details.
As a scoring sanity check, we collected community reference solutions from online NCRE study communities for all 200 tasks.
These solutions, created informally by exam candidates with no guarantee of correctness, achieve an average score of 95.5\% (89 out of 200 perfect scores). This indicates that high-scoring solutions exist for the tasks and that the scoring engine produces expected results on well-formed documents.
Manual inspection of low-scoring community reference solutions suggests that much of the remaining gap comes from imperfections in the community answers themselves (see Appendix~\ref{sec:community_errors}).

\subsection{Taxonomy of Office skills}
\label{sec:taxonomy}

Based on the criterion type annotations in the scoring configurations, we organize \name's 7,118 scoring criteria into 8 skill categories that span all three applications.
Table~\ref{tab:taxonomy} presents this taxonomy with the distribution of criteria across the full dataset.

\begin{table}[t]
  \caption{Taxonomy of Office skills in \name. All 7,118 scoring criteria across 200 tasks are organized into 8 categories based on their criterion types. The ``Apps'' column indicates which applications contribute criteria to each category.}
  \label{tab:taxonomy}
  \centering
  \small
  \begin{tabular}{lp{4.2cm}lrr}
    \toprule
    \textbf{Category} & \textbf{Description} & \textbf{Apps} & \textbf{Count} & \textbf{\%} \\
    \midrule
    Text \& Format & Paragraph styles, fonts, text content, find/replace & W, E, P & 2,744 & 38.6 \\
    Layout \& Design & Page setup, margins, slide layout, themes, masters & W, E, P & 1,253 & 17.6 \\
    Tables & Table creation, structure, cell formatting & W, E, P & 771 & 10.8 \\
    Graphics \& Media & Images, SmartArt, shapes, WordArt & W, P & 649 & 9.1 \\
    Doc.\ Properties & Document settings, worksheet management & W, E, P & 613 & 8.6 \\
    Animation & Slide animations and transition effects & P & 555 & 7.8 \\
    Charts & Chart type, axes, legends, data series & W, E, P & 362 & 5.1 \\
    Data \& Formulas & Formulas, pivot tables, conditional formatting & E & 171 & 2.4 \\
    \midrule
    \textbf{Total} & & & \textbf{7,118} & \\
    \bottomrule
  \end{tabular}
\end{table}

This taxonomy enables per-skill capability analysis beyond aggregate scores (Section~\ref{sec:skill_analysis}).

\subsection{Dataset statistics}
\label{sec:statistics}

Table~\ref{tab:statistics} presents the overall statistics of \name.
The benchmark comprises 200 tasks across two difficulty levels with a total of 7,118 scoring criteria, providing dense evaluation coverage.

\begin{table}[t]
  \caption{Dataset statistics of \name. Each task is scored on a 100-point scale with fine-grained criteria. Level~1 tasks assess foundational skills; Level~2 tasks cover advanced operations.}
  \label{tab:statistics}
  \centering
  \small
  \begin{tabular}{lrrrr}
    \toprule
    & \textbf{Word} & \textbf{Excel} & \textbf{PPT} & \textbf{Total} \\
    \midrule
    \multicolumn{5}{l}{\textit{Level~1 (Foundational)}} \\
    Tasks & 28 & 28 & 32 & 88 \\
    Scoring criteria & 1,604 & 638 & 1,425 & 3,667 \\
    Avg.\ criteria per task & 57.3 & 22.8 & 44.5 & 41.7 \\
    \midrule
    \multicolumn{5}{l}{\textit{Level~2 (Advanced)}} \\
    Tasks & 38 & 37 & 37 & 112 \\
    Scoring criteria & 1,090 & 1,236 & 1,125 & 3,451 \\
    Avg.\ criteria per task & 28.7 & 33.4 & 30.4 & 30.8 \\
    \midrule
    \multicolumn{5}{l}{\textit{Combined}} \\
    Tasks & 66 & 65 & 69 & 200 \\
    Scoring criteria & 2,694 & 1,874 & 2,550 & 7,118 \\
    Max score per task & 100 & 100 & 100 & 100 \\
    \bottomrule
  \end{tabular}
\end{table}

Level~1 tasks have more criteria per task (41.7 avg.) with simpler checks, while Level~2 tasks have fewer (30.8 avg.) but more complex compound rules.
With 7,118 independent evaluation signals across 200 tasks, the evaluation density is substantially higher than benchmarks using binary pass/fail.

\section{Experimental setup}
\label{sec:experiments}

\subsection{Models}
\label{sec:models}

We evaluate 7 frontier LLMs: four proprietary/API systems (Claude Opus 4.7~\citep{anthropic2026opus47}, GPT-5.5~\citep{openai2026gpt55}, Gemini 3.1 Pro~\citep{google2026gemini31pro}, Grok-4.1-fast~\citep{xai2025grok41fast}) and three open-weight models (Kimi-K2.6~\citep{moonshot2026kimi26}, Qwen3.5-397B-A17B~\citep{qwen2026qwen35}, MiMo-V2.5~\citep{xiaomi2026mimov25}).
Because task instructions often contain reference images (Section~\ref{sec:task_format}), all models must support multimodal input, which excludes competitive text-only LLMs such as DeepSeek-R1~\citep{deepseek2025r1} and GPT-5.3-Codex-Spark~\citep{openai2026codexspark}.
For the open-weight group, we selected three multimodal models available through our evaluation infrastructure at the time of evaluation to maximize the chance of observing high Office performance.

\subsection{Agent architecture}
\label{sec:agent}

We evaluate code-based methods in two settings: a \emph{single-turn LLM baseline} that probes the underlying model directly with a single API call, and \emph{autonomous coding agents} that exercise the model within an iterative scaffold~\citep{wang2024executable}. In both settings the model produces executable code that manipulates documents, rather than interacting with a GUI.

\paragraph{Single-turn LLM baseline.}
This is a single API call with no agent loop. The model receives pre-processed task instructions, document screenshots, and the input document path, and produces Python code in one pass using standard Office libraries (\texttt{python-docx}, \texttt{openpyxl}, \texttt{python-pptx}). There is no execution feedback, no retry, and no tool use beyond emitting code. This setting probes single-turn code generation under a fixed library-constrained automation interface.

\paragraph{Coding agent.}
We evaluate two autonomous agents: Claude Code (CC), powered by Claude Opus 4.7~\citep{anthropic2026opus47}, and Codex, powered by GPT-5.5~\citep{openai2026gpt55}. Each iteratively writes, executes, and debugs code with no turn limit and unrestricted tool access.
They receive only the raw exam paper and input documents, the same materials given to a human candidate sitting the NCRE exam, without pre-processed descriptions, screenshots, or scoring rubrics.
This setting is a stronger system comparison, not an ablation of a single factor. It differs from the single-turn baseline in execution feedback, repair budget, agent scaffolding, and access to Office automation tools such as COM.

\subsection{Evaluation metrics}
\label{sec:metrics}

Our primary metric is \textbf{Score Rate (SR)}: the unweighted macro-average score across the 200 extracted practical-operation tasks, expressed as a percentage of the 100-point maximum:
$\text{SR} = \frac{1}{N}\sum_{i=1}^{N} s_i\;\%$,
where $s_i$ is the score on task $i$.
Since each task is scored on a 100-point scale and contributes equally to the average, SR is numerically equal to the mean task score.
Application-, level-, and skill-specific scores are reported as analytical breakdowns; they are not reweighted to match the full NCRE exam composition, and SR remains the single metric used for overall model comparison.

\section{Results}
\label{sec:results}

\subsection{Main results}
\label{sec:main_results}

Table~\ref{tab:main_results} presents our main results.
The best single-turn model, Claude Opus 4.7, reaches only 36.6\%, followed by GPT-5.5 at 36.2\%. Both are far below the 60-point reference threshold used to contextualize this extracted practical subset and below the 95.5\% community-reference score (Section~\ref{sec:eval}).
All numbers are means across 3 independent runs (SD 0.4--2.1pp on overall SR).
A paired $t$-test across 200 tasks confirms that the top two models, Claude Opus 4.7 and GPT-5.5, are not statistically distinguishable ($p{=}0.82$; bootstrap 95\% CI for the difference: $[-3.5, 4.4]$pp).

\begin{table}[t]
  \caption{Main results on \name. SR\%: Score Rate per application and level (mean of 3 runs). Exec\%: fraction of tasks where the generated code runs without error. L1 = Level~1; L2 = Level~2. Bold = best per column.}
  \label{tab:main_results}
  \centering
  \small
  \setlength{\tabcolsep}{4pt}
  \begin{tabular}{l cc cc cc c c}
    \toprule
    & \multicolumn{2}{c}{\textbf{Word}} & \multicolumn{2}{c}{\textbf{Excel}} & \multicolumn{2}{c}{\textbf{PPT}} & & \\
    \cmidrule(lr){2-3} \cmidrule(lr){4-5} \cmidrule(lr){6-7}
    \textbf{Model} & L1 & L2 & L1 & L2 & L1 & L2 & \textbf{Overall} & \textbf{Exec\%} \\
    \midrule
    \multicolumn{9}{l}{\textit{Proprietary/API}} \\
    Claude Opus 4.7 & \textbf{58.5} & \textbf{29.1} & 46.4 & 43.9 & 21.8 & \textbf{25.9} & \textbf{36.6} & \textbf{61.5} \\
    GPT-5.5 & 34.8 & 26.8 & \textbf{61.7} & \textbf{51.6} & \textbf{28.7} & 18.4 & 36.2 & 56.8 \\
    Gemini 3.1 Pro & 14.5 & 8.8 & 39.7 & 22.8 & 8.3 & 8.1 & 16.3 & 29.4 \\
    Grok-4.1-fast & 2.2 & 2.8 & 6.3 & 1.7 & 1.1 & 3.0 & 2.8 & 2.2 \\
    \midrule
    \multicolumn{9}{l}{\textit{Open-weight}} \\
    Kimi-K2.6 & 7.6 & 4.7 & 28.5 & 12.9 & 1.5 & 4.6 & 9.4 & 19.2 \\
    Qwen3.5-397B-A17B & 4.5 & 4.3 & 7.1 & 3.8 & 1.0 & 3.4 & 3.9 & 15.0 \\
    MiMo-V2.5 & 4.5 & 3.5 & 5.7 & 3.0 & 1.3 & 3.6 & 3.5 & 9.8 \\
    \bottomrule
  \end{tabular}
\end{table}

Key observations:
\begin{itemize}[nosep, leftmargin=*]
    \item \textbf{All single-turn models remain weak on the extracted practical subset.} Even on the strongest category (Excel L1, GPT-5.5: 61.7\% SR), nearly 40\% of task score is lost. Overall SR ranges from 36.6\% to 2.8\%, and the proprietary/API group contains both the top systems and the weakest system (Grok-4.1-fast). Most models degrade substantially on Level~2, with drops more pronounced for stronger models.
    \item \textbf{Under the single-turn library-constrained setting, Excel is the most tractable and PowerPoint the hardest.} Excel tasks are dominated by formulas and tabular data manipulation, where the operation maps cleanly to a small, well-known set of programmatic primitives. PowerPoint tasks more often hinge on animation effects, transitions, and theme/master constants whose exact names and parameters are not stated in the instruction and must be recalled from the model's internal knowledge of the Office object model.
    \item \textbf{Code reliability is a major bottleneck.} The best model's Exec\% is only 61.5 (Table~\ref{tab:main_results}), meaning the remaining 38.5\% of programs crash before producing output. Among the tasks that do run, the implied conditional SR is roughly 60\% (overall SR divided by Exec\%), showing that accuracy remains a major problem even when code executes.
\end{itemize}

\subsection{Skill-level analysis}
\label{sec:skill_analysis}

\begin{figure}[t]
  \centering
  \includegraphics[width=0.75\linewidth]{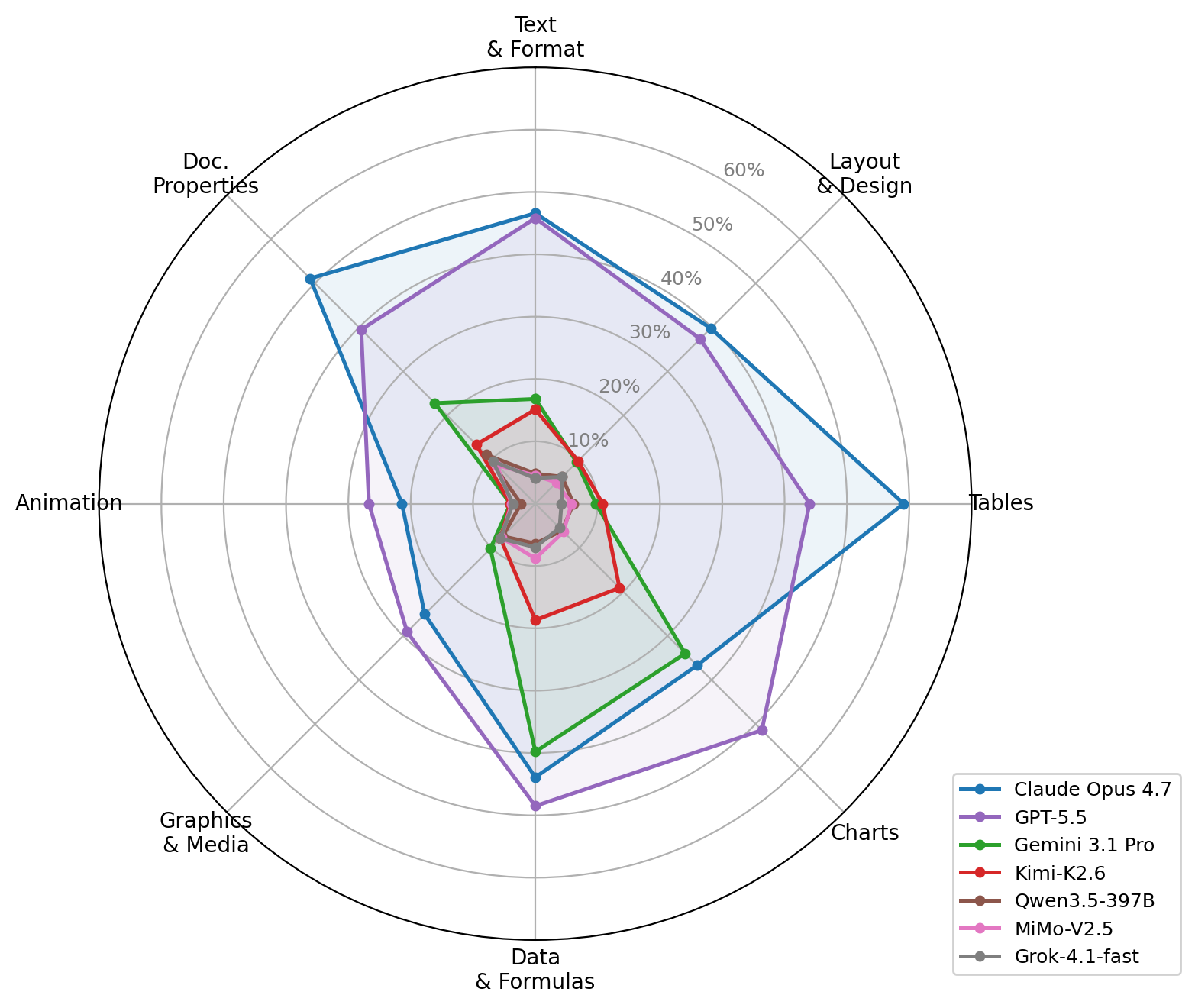}
  \caption{Per-skill criterion pass rate (\%) for all seven evaluated models. Tables, Charts, and Data \& Formulas contain the strongest peaks among top models; Animation and Graphics \& Media are the weakest. See Table~\ref{tab:perskill_full} for full results.}
  \label{fig:radar}
\end{figure}

Figure~\ref{fig:radar} shows the per-skill criterion pass rate for all seven evaluated models. Models exhibit distinct capability profiles across the 8 skill categories.

Full per-skill results appear in Table~\ref{tab:perskill_full} (Appendix). Key patterns:

\begin{itemize}[nosep, leftmargin=*]
    \item \textbf{Structured, explicitly specified operations are strongest.} Tables (59.1\% for Claude Opus 4.7), Charts (51.4\% for GPT-5.5), and Data \& Formulas (48.5\% for GPT-5.5) reach the highest category-level pass rates. These categories rely more on explicit task parameters and well-scoped document structures, reducing the implementation knowledge gap.
    \item \textbf{Animation and Graphics \& Media are weakest} (26.7\% and 29.1\% at best). These require exact enumeration names and visual property values absent from the task text; the implementation knowledge gap is widest here.
    \item \textbf{Weaker models show flat profiles}: open-weight models' pass rates vary little across categories (e.g., Qwen3.5-397B-A17B: 2.3--11.1\%), suggesting failure at a more basic level before category-specific challenges become relevant.
\end{itemize}

The skill profiles also help explain the per-application gains from coding agents (Table~\ref{tab:coding_agent}).
PowerPoint sees the largest agent improvement (Codex: +43.8pp over single-turn GPT-5.5), which aligns with Animation and Graphics~\&~Media being the weakest single-turn categories: these skills rely heavily on exact enumeration constants that \texttt{python-pptx} often lacks, but that COM automation exposes directly.
Excel gains are also large (+26.3pp), consistent with Data~\&~Formulas and Charts already being the strongest single-turn categories; the stronger agentic setting recovers from many crashes that suppress single-turn scores.
Word gains are the smallest for Claude Code (+0.7pp) despite Word's Text~\&~Format category being reasonably strong (46.6\%); this suggests that Word's bottleneck is implementation knowledge (exact font names, style IDs, page-setup constants) rather than crashes, a bottleneck that iterative repair alone cannot address.

\subsection{Coding agent results}
\label{sec:coding_agent}

To measure a stronger autonomous setting, we evaluate two coding agents: Claude Code (CC), powered by Claude Opus 4.7, and Codex, powered by GPT-5.5. This setting combines execution feedback, iterative repair, agent scaffolding, and unrestricted tool access, with a one-hour time limit per task. The agents receive only the raw exam paper and input documents, without pre-processed descriptions, screenshots, or scoring rubrics. The reported turn counts (Table~\ref{tab:coding_agent}) reflect natural convergence within the time limit, not an imposed turn cap.

\begin{table}[t]
  \caption{System-level comparison between the single-turn LLM baseline and autonomous coding agents. SR\% is reported per application (averaged across both levels) and overall; Exec\% is the code execution success rate; Turns is the average number of agent interaction rounds.}
  \label{tab:coding_agent}
  \centering
  \small
  \begin{tabular}{l cccc cc}
    \toprule
    \textbf{Agent} & \textbf{Word} & \textbf{Excel} & \textbf{PPT} & \textbf{Overall} & \textbf{Exec (\%)} & \textbf{Turns} \\
    \midrule
    Single-turn (GPT-5.5) & 30.2 & 55.9 & 23.2 & 36.2 & 56.8 & 1 \\
    Single-turn (Claude Opus 4.7) & 41.6 & 44.9 & 24.0 & 36.6 & 61.5 & 1 \\
    Coding agent (CC, Claude Opus 4.7) & 42.3 & 66.3 & 50.7 & 53.0 & 98.0 & 23.1 {\scriptsize(3--90)} \\
    Coding agent (Codex, GPT-5.5) & \textbf{57.4} & \textbf{82.2} & \textbf{67.0} & \textbf{68.8} & \textbf{99.0} & 50.0 {\scriptsize(8--133)} \\
    \bottomrule
  \end{tabular}
\end{table}

The best coding agent (Codex, GPT-5.5) achieves 68.8\% overall SR, a 32.6pp improvement over its single-turn counterpart. Claude Code (Claude Opus 4.7) reaches 53.0\%, a 16.4pp gain over its single-turn baseline. Both agents close a substantial portion of the gap to the community-reference score (95.5\%), but neither closes it fully. In this stronger agentic setting, execution success rises to 98--99\% for both agents. Both agents also exercise the broader tool access of this setting. For example, they drive Office through \texttt{win32com} COM automation in addition to the standard libraries, which is particularly impactful for PowerPoint operations.
These results should be read as a comparison between two system settings, not as proof that iterative debugging alone causes the improvement. The coding-agent setting changes several things at once: execution feedback, repair budget, agent scaffolding, and the range of tools the model is allowed to invoke. The single-turn baseline, by design, exercises only the libraries provided in its prompt; separating the effects of feedback, broader tool access, and agent scaffolding remains future work.
Despite these gains, the best coding agent still remains below the community-reference score (95.5\%). Once many code-execution failures are fixed, models still lose substantial credit on fine-grained Office details, such as exact style names, color values, chart layouts, animation constants, and XML properties.

\section{Error analysis}
\label{sec:error_analysis}

We analyze failure modes of the best single-turn model (Claude Opus 4.7) and categorize common issues:
\textbf{(1)~Code execution failure}: hallucinated API methods or type errors crash the program.
\textbf{(2)~Missing or misunderstood operation}: required operations are skipped, misunderstood, or only partially completed (e.g., formatting only the first paragraph).
\textbf{(3)~Implementation-knowledge error}: the generated code targets the right high-level operation but uses the wrong API parameter, built-in style name, color value, chart layout, animation constant, or XML property.
\textbf{(4)~Cascading failure}: an upstream missing object or incorrect transformation causes downstream criteria to fail.
\textbf{(5)~Library limitation}: the intended operation is correct but unsupported by the selected library (e.g., certain PowerPoint animations, advanced mail merge).

Code execution failure accounts for most zero-score outcomes (23.7\% of tasks averaged across three runs), disproportionately affecting PPT (29.5\%) and Excel (26.7\%) over Word (14.6\%).
To understand the errors beyond crashes, we annotate failed scoring criteria for Claude Opus 4.7 single-turn (run~1) and the Codex coding agent. For Claude Opus 4.7, we use execution status and generated-code keyword matching to separate operations that were attempted but wrong from operations that were missing. For Codex, the classification relies mainly on scorer errors and final outputs. When the target object exists but the exact property is wrong, we label the error as implementation knowledge. This is a diagnostic labeling scheme, not a definitive manual audit; it may over-count implementation-knowledge errors for Codex and under-count library limitations when unsupported operations are simply skipped.

Table~\ref{tab:error_taxonomy} suggests a shift in the error profile under our diagnostic labeling scheme. The table uses weighted failure units from the error-taxonomy script, not the 200-task $\times$ 100-point SR denominator, so the percentages should be read only as a breakdown of annotated errors within each setting. In the single-turn setting, execution failures account for 51.8\% of all weighted loss; among non-crash losses, implementation-knowledge errors are already the largest category (61.7\%). In the coding-agent setting, execution failures drop to 7.9\% of all weighted loss, while 97.4\% of non-crash weighted loss is labeled as implementation knowledge. The larger weighted implementation-knowledge loss for the coding agent (1,336.8 vs.\ 859.5) should not be read as degradation: after crashes are fixed, many fine-grained property errors become visible instead of being hidden by execution failure.

\begin{table}[t]
  \caption{Failure taxonomy over failed scoring criteria. W. loss denotes weighted loss computed from criterion-level annotation weights and is used only for relative error decomposition within this table; it is not the task-normalized SR denominator. ``All loss'' uses all failed criteria in the setting; ``Non-crash loss'' excludes execution-failure rows.}
  \label{tab:error_taxonomy}
  \centering
  \small
  \setlength{\tabcolsep}{4pt}
  \begin{tabular}{llrrrr}
    \toprule
    \textbf{Setting} & \textbf{Failure type} & \textbf{Criteria} & \textbf{W. loss} & \textbf{All loss \%} & \textbf{Non-crash loss \%} \\
    \midrule
    \multirow{5}{*}{Single-turn Opus} & Execution failure & 2,314 & 1,494.5 & 51.8 & -- \\
    & Implementation knowledge & 1,183 & 859.5 & 29.8 & 61.7 \\
    & Missing/misunderstood & 700 & 500.9 & 17.3 & 36.0 \\
    & Cascading failure & 23 & 26.0 & 0.9 & 1.9 \\
    & Library limitation & 9 & 6.8 & 0.2 & 0.5 \\
    \midrule
    \multirow{3}{*}{Codex coding agent} & Execution failure & 119 & 118.4 & 7.9 & -- \\
    & Implementation knowledge & 1,868 & 1,336.8 & 89.7 & 97.4 \\
    & Missing/misunderstood & 29 & 35.2 & 2.4 & 2.6 \\
    \bottomrule
  \end{tabular}
\end{table}

\begin{figure}[t]
  \centering
  \includegraphics[width=\linewidth]{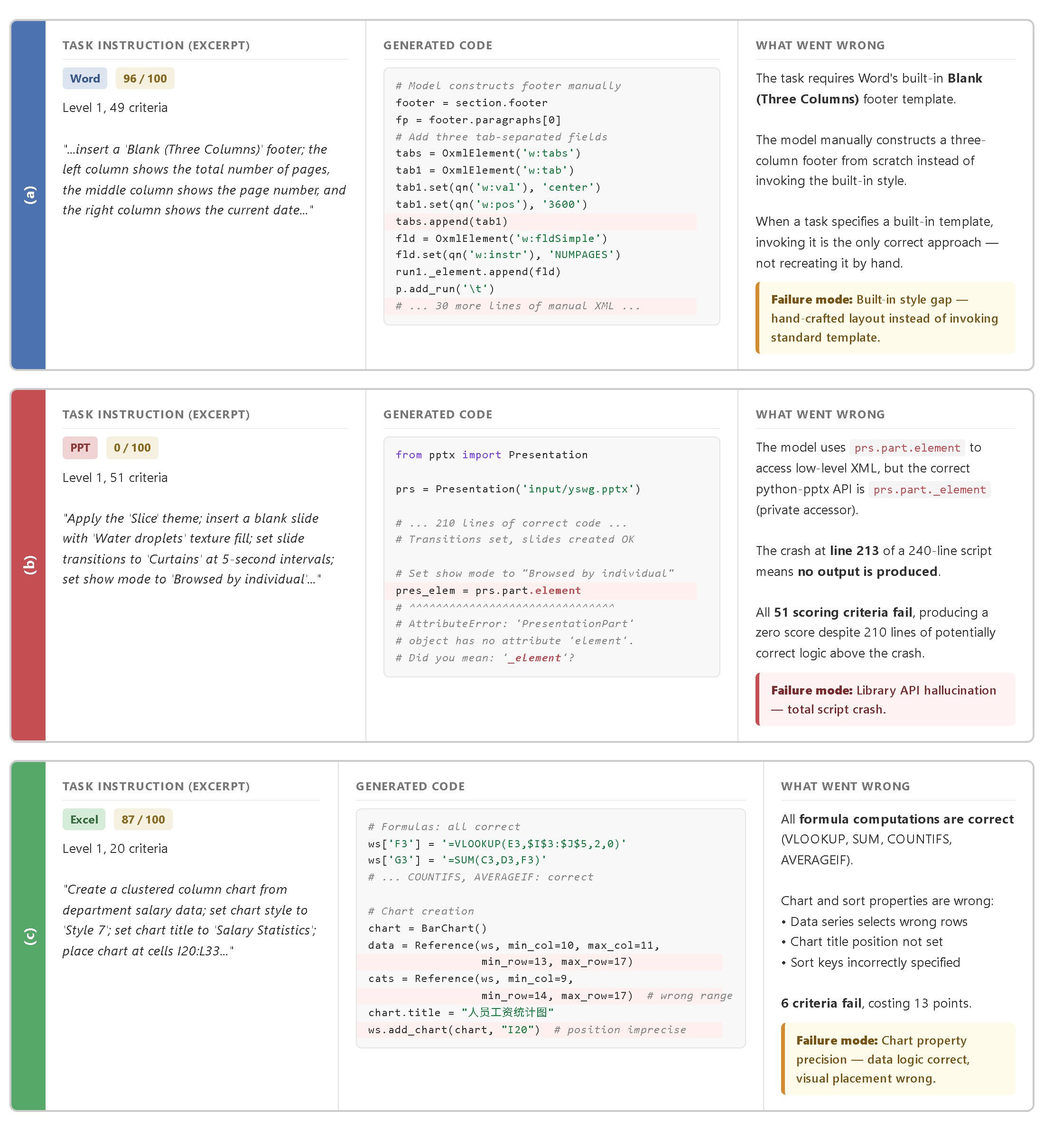}
  \caption{Representative failure cases from \name, using Claude Opus 4.7 outputs. (a)~A Word task where the model reconstructs a footer manually instead of using Word's built-in template. (b)~A zero-score PPT task where the model hallucinates a non-existent API, crashing the entire script. (c)~An Excel task where all formulas are correct but chart properties are wrong, costing 13 points.}
  \label{fig:error_cases}
\end{figure}

Table~\ref{tab:ik_subtypes} breaks down implementation-knowledge errors. The dominant subtypes are not semantic misunderstandings of the instructions, but low-level representation mismatches: Open XML property paths, enumeration constants, color/theme encodings, numeric units, SmartArt/shape properties, and chart layouts. Built-in template mismatches are a recurring example: tasks requiring named templates (e.g., ``Blank (Three Columns)'' footer) are reconstructed from scratch instead of invoked by name. Formula cascading failures and chart property imprecision further reduce scores on otherwise strong Excel tasks (see Figure~\ref{fig:error_cases} and Appendix for detailed examples).

\begin{table}[t]
  \caption{Dominant implementation-knowledge subtypes. W. loss denotes weighted loss. Shares are within implementation-knowledge weighted loss for each setting; rows show the six largest subtypes by Codex-agent weighted loss.}
  \label{tab:ik_subtypes}
  \centering
  \small
  \setlength{\tabcolsep}{5pt}
  \begin{tabular}{lrrr rrr}
    \toprule
    & \multicolumn{3}{c}{\textbf{Single-turn Opus}} & \multicolumn{3}{c}{\textbf{Codex coding agent}} \\
    \cmidrule(lr){2-4} \cmidrule(lr){5-7}
    \textbf{Subtype} & \textbf{Criteria} & \textbf{W. loss} & \textbf{\%} & \textbf{Criteria} & \textbf{W. loss} & \textbf{\%} \\
    \midrule
    OOXML property/path & 391 & 289.7 & 33.7 & 413 & 344.9 & 25.8 \\
    Enumeration/constant & 239 & 179.9 & 20.9 & 432 & 299.5 & 22.4 \\
    Color/theme/gradient & 170 & 116.6 & 13.6 & 275 & 197.4 & 14.8 \\
    Numeric value/unit & 172 & 116.1 & 13.5 & 233 & 157.8 & 11.8 \\
    SmartArt/shape & 37 & 37.7 & 4.4 & 184 & 119.9 & 9.0 \\
    Chart layout/style & 59 & 38.8 & 4.5 & 90 & 75.3 & 5.6 \\
    \bottomrule
  \end{tabular}
\end{table}

\paragraph{Coding agent failure modes.}
The coding-agent setting greatly reduces crashes (98--99\% execution success vs.\ 57--62\% single-turn), but introduces iterative regression: the best agent (Codex) scores worse than its single-turn counterpart on 23 tasks (11.5\%), sometimes corrupting working documents through successive modifications.
This trade-off is inherent to repair-based workflows: execution-time correction can recover crashed tasks but risks undoing correct work.

\section{Discussion and conclusion}
\label{sec:discussion}

\paragraph{Single-turn vs.\ agentic systems.}
The gap between single-turn (36.6\%) and coding agents (68.8\%) maps onto a design choice in real AI-assisted workflows. Single-turn code generation is a one-shot assistance paradigm: the AI produces output in one pass. Stronger agentic systems attempt autonomous task completion through execution feedback, repair budget, scaffolding, and broader tool access. The one-shot setting fails on long-horizon Office tasks; most generated programs crash without producing output. The stronger agentic setting recovers from many of these crashes, but at the cost of regression on already-correct work. For deployment, reliable Office automation likely needs closed-loop execution with safeguards against regression and careful control over tool access.

\paragraph{Code generation as a proxy.}
Programmatic automation is a powerful and scalable proxy for many Office operations, especially when COM exposes the relevant object-model controls. However, GUI-based agents with visual feedback may access UI-level affordances, galleries, theme assets, and layout heuristics that are difficult to reproduce through library calls or COM constants alone. A direct comparison between programmatic and GUI-based Office agents remains an important direction for future work.

\paragraph{Why Office tasks are hard for AI.}
Our analysis shows qualitative differences between human and AI approaches:
\textbf{(1)~Style galleries vs.\ API constants.} Humans select built-in templates and effects from visual galleries in 2--3 clicks; AI agents must produce exact internal constants (color hex values, style IDs, enumeration names) \emph{nowhere in the task description}. Our criterion-level taxonomy shows that this implementation-knowledge gap remains a persistent bottleneck after many crashes are repaired.
\textbf{(2)~Limited interactive visual feedback.} Humans receive continuous visual feedback while editing. The evaluated systems receive either static screenshots in the single-turn setting or raw documents in the coding-agent setting, but they do not receive the same continuous, closed-loop visual confirmation after each operation. This helps explain why visually specified categories (Animation: 26.7\%, Graphics \& Media: 29.1\%) lag far behind more structured categories with explicit parameters (Tables: 59.1\%, Charts: 51.4\%, Data \& Formulas: 48.5\%).
\textbf{(3)~Cascading failures.} In multi-step tasks, if an early operation fails, all downstream operations score zero. Humans avoid this through sequential visual confirmation.
\textbf{(4)~Iterative regression.} Coding agents can \emph{score worse} than single-turn on some tasks: iterative debugging sometimes corrupts previously correct output, a failure mode absent in human workflows.

\paragraph{Skill-augmented agents.}
Recent work on skill-based agent architectures~\citep{huang2026skillsbench,li2026skillagentswild} shows that equipping agents with curated, reusable skill libraries can substantially improve task completion.
Our error analysis provides a concrete case for this direction in the Office domain.
Table~\ref{tab:error_taxonomy} shows that, under this diagnostic labeling scheme, 97.4\% of non-crash weighted loss for the best coding agent is labeled as implementation knowledge, and Table~\ref{tab:ik_subtypes} further breaks this down into actionable subtypes: OOXML property paths (25.8\%), enumeration constants (22.4\%), color/theme encodings (14.8\%), and numeric units (11.8\%).
These are the kinds of errors that pre-defined skills may help address.
For instance, a skill for ``set slide transition'' could encapsulate the exact \texttt{ppTransitionType} enumeration value, the effect-option mapping, and the duration unit conversion---knowledge that current models must guess from memory.
Per-skill pass rates (Table~\ref{tab:perskill_full}) suggest where such skills would yield the highest return: Animation (best model: 26.7\%) and Graphics~\&~Media (29.1\%) are the weakest categories, while Tables (59.1\%) and Data~\&~Formulas (48.5\%) are already comparatively strong.
Designing and evaluating a skill library for Office automation, and measuring how much of the implementation-knowledge bottleneck it can close, is a promising direction for future work.

\paragraph{Cross-language analysis.}
A natural question is whether the performance gap stems primarily from the difficulty of the Office tasks themselves or from models' weaker handling of Chinese-language task materials. To investigate, we construct an English-translated variant of all 200 tasks, including document content, scoring criteria, and font/style mappings, and re-evaluate 5 models under the same single-turn setting. Table~\ref{tab:crosslang} reports the results.

\begin{table}[t]
  \caption{Cross-language comparison (single-turn SR\%). EN = English-translated tasks; ZH = original Chinese tasks. Claude Opus~4.7 EN is the mean of 2 runs. Grok-4.1-fast and Kimi-K2.6 are omitted due to API unavailability at the time of the English experiment.}
  \label{tab:crosslang}
  \centering
  \small
  \setlength{\tabcolsep}{4pt}
  \begin{tabular}{l cc cc cc cc}
    \toprule
    & \multicolumn{2}{c}{\textbf{Word}} & \multicolumn{2}{c}{\textbf{Excel}} & \multicolumn{2}{c}{\textbf{PPT}} & \multicolumn{2}{c}{\textbf{Overall}} \\
    \cmidrule(lr){2-3} \cmidrule(lr){4-5} \cmidrule(lr){6-7} \cmidrule(lr){8-9}
    \textbf{Model} & ZH & EN & ZH & EN & ZH & EN & ZH & EN \\
    \midrule
    Claude Opus 4.7 & \textbf{41.6} & 33.3 & \textbf{45.0} & 37.4 & 24.0 & \textbf{33.6} & \textbf{36.6} & 34.7 \\
    GPT-5.5 & \textbf{30.2} & 30.3 & \textbf{56.0} & 48.9 & 23.2 & \textbf{40.8} & 36.2 & \textbf{40.0} \\
    Gemini 3.1 Pro & 11.2 & \textbf{18.7} & 30.1 & \textbf{27.4} & 8.2 & \textbf{21.1} & 16.3 & \textbf{22.4} \\
    Qwen3.5-397B & 4.4 & \textbf{12.3} & 5.2 & \textbf{7.7} & 2.3 & \textbf{17.7} & 3.9 & \textbf{12.7} \\
    MiMo-V2.5 & 3.9 & \textbf{12.0} & 4.2 & 3.0 & 2.5 & \textbf{19.9} & 3.5 & \textbf{11.8} \\
    \bottomrule
  \end{tabular}
\end{table}

Three patterns emerge.
\textbf{(1)~The translated variant does not remove the difficulty for strong models.} Claude Opus~4.7 performs comparably in both languages (34.7\% EN vs.\ 36.6\% ZH, $\Delta{=}{-}1.9$pp), and GPT-5.5 slightly improves (+3.8pp). This suggests that the best models' failures are driven substantially by Office-specific implementation knowledge, not only by Chinese instruction comprehension.
\textbf{(2)~Mid-tier and open-weight models benefit substantially from English.} Gemini~3.1~Pro improves by +6.1pp, Qwen3.5 by +8.8pp, and MiMo by +8.3pp when switching to English, indicating that Chinese instruction understanding is a meaningful secondary bottleneck for these models.
\textbf{(3)~PowerPoint shows the largest translated-variant gains.} Nearly all models gain 10--17pp on PPT in English. PPT operations reference named constants (e.g., ``Emphasis/Teeter,'' ``Medium Style~2'') that originate from English-language Office APIs; English translations and font/style mappings align more directly with these identifiers.
Crucially, even on English tasks, the best single-turn model reaches only 40.0\% (GPT-5.5), still well below the 60-point reference threshold for the extracted practical subset. This suggests that English translation can affect some models and applications, especially PowerPoint, but the task difficulty remains substantial in the translated variant.

\paragraph{Limitations.}
\textbf{(1)} The English experiment (Table~\ref{tab:crosslang}) uses translated NCRE tasks rather than a natively English certification (e.g., MOS). Translation artifacts---such as font-name substitutions and style-constant mappings---may introduce confounds that a native English exam would avoid.
\textbf{(2)} We evaluate code-generation agents only; GUI-based agents with visual feedback might perform differently, particularly on layout-heavy tasks. A direct comparison remains future work.
\textbf{(3)} The coding-agent comparison changes several factors at once: execution feedback, repair budget, agent scaffolding, and COM access. Future ablations should isolate these factors.
\textbf{(4)} Community reference solutions (95.5\%) provide a useful scoring sanity check but are not a formal human baseline. NCRE materials may also appear in public training data; the direction and magnitude of possible contamination effects are difficult to determine.
\textbf{(5)} NCRE materials remain copyrighted and are not redistributed by the authors. Exact replication therefore requires researchers to obtain the same materials independently. At the same time, this is what gives the study its exam anchor: the tasks were written and calibrated outside this work. We document the selection protocol, task statistics, scoring taxonomy, evaluation pipeline, prompts, and criterion-level aggregate results to make the study as inspectable as possible (Section~\ref{sec:data_source}).

\paragraph{Conclusion.}
On this extracted practical-operation subset of a standardized Office proficiency exam, no single-turn LLM earns more than 36.6\% of the officially allocated points. A stronger autonomous coding-agent setting reaches 53.0--68.8\%, but remains below the 95.5\% community-reference score. \name uses a real certification exam to track how close code-generating LLM and agent systems are to reliable, fine-grained Office document automation.

\bigskip
\noindent\textbf{Authors.}\quad
Tengchao Lv,
Dongdong Zhang,
Jiayu Ding,
Yilin Jia,
Yuzhong Zhao,
Yupan Huang,
Wenshan Wu,
Xiangyang Zhou,
Shaohan Huang,
Nan Yang,
Li Dong,
Lei Cui,
Furu Wei

\bibliographystyle{plainnat}
\bibliography{references}


\newpage
\appendix

\section{Task examples}
\label{app:examples}

We provide three complete task examples (one per application) illustrating the format, difficulty, and scoring granularity of \name tasks. All three are Level~1 (foundational) tasks. The original instructions are in Chinese; we provide English translations below each.

\subsection{Word task: First China Online Media Forum (71 criteria)}

\begin{CJK*}{UTF8}{gbsn}
\small
\begin{quote}
在考生文件夹下，打开文档WORD.docx，按照要求完成下列操作并以该文件名（WORD.docx）保存文档。

⑴将文中所有错词"网罗"替换为"网络"；将标题段文字（"首届中国网络媒体论坛在青岛开幕"）的字体格式设置为三号黑体、加粗、居中、字符宽度调整为17字符，文本渐变填充效果设为"预设渐变/径向渐变 - 个性色2,类型/路径,颜色/红色（标准色）"，文本映像效果设为"预设/映像变体/全映像，4pt偏移量：透明度50\%、大小65\%、距离2.5磅"。

⑵设置页面纸张大小为"16开（18.4厘米 x 26厘米）"；在页面底端插入"椭圆形"样式页码，页脚底端距离2厘米，设置页码编号格式为"壹，贰，叁，……"、起始页码为"贰"；编辑文档属性信息：标题/首届中国网络媒体论坛在青岛开幕，作者/A考生，单位/NCRE；在页面顶端插入"空白"型页眉，页眉内容为该文档作者（"文档部件/文档属性"）；将页面颜色的填充效果设置为"纹理/羊皮纸"；为页面添加内容为"首届中国网络媒体论坛"文字型水印，水印内容的文本格式为：黑体、蓝色（标准色）。

⑶将正文各段文字（"6月22日，……评选办法等。"）设置为12磅方正姚体；第一段首字下沉2行，距正文0.2厘米；除第一段外的其余各段落（不包括表格）左、右各缩进1.5字符，首行缩进2字符，段前间距1行；将正文第三段（"论坛的主题是……管理和自律。"）分为等宽两栏，设置栏宽为15字符，栏间加分隔线；将正文第四段（"与会嘉宾……评选办法等。"）的段落分页设置为"段中不分页"。

⑷在表格顶端添加表标题"公众号关注量统计表"，并将其设置为小二号华文彩云、加粗、居中、深红（标准色）；在表格右侧插入一空列，在该列第一行的单元格中输入列标题"合计"，其余各单元格中填入该行各单元格数据的总和（利用表格工具中的公式），按"合计"列依据"数字"类型对表格降序排序；设置表格居中，表格第一行和第一列内容居中、垂直对齐为水平居中，其余单元格内容右对齐；设置表格行高为0.6厘米、第一列列宽为2.5厘米、其余列列宽1.8厘米。

⑸设置表格第一行底纹颜色为主题颜色"水绿色，个性色5，淡色80\%"；设置表格外框线和第一行的下框线为红色（标准色）0.75磅双窄线、其余内框线为红色（标准色）0.5磅单实线。
\end{quote}

\normalsize
\noindent\textbf{English translation:}
\small
\begin{quote}
Open WORD.docx and complete the following operations:

1. Replace all instances of the typo ``网罗'' with ``网络''; format the title (``First China Online Media Forum Opens in Qingdao''): SimHei 16pt, bold, centered, character width 17; set text gradient fill to ``Preset Gradient/Radial -- Accent 2, Type/Path, Color/Red (standard)''; text reflection to ``Full Reflection, 4pt offset: transparency 50\%, size 65\%, distance 2.5pt.''

2. Set page size to 16K (18.4cm $\times$ 26cm); insert ``Oval'' page number at bottom, footer distance 2cm, numbering format ``壹, 贰, 叁, \ldots'' starting at ``贰''; set document properties: title/First China Online Media Forum Opens in Qingdao, author/A考生, organization/NCRE; insert ``Blank'' header with document author (via Document Properties); set page color fill to ``Texture/Parchment''; add text watermark ``首届中国网络媒体论坛'' in SimHei, blue (standard).

3. Format body paragraphs (``6月22日, \ldots 评选办法等.''): 12pt FangZheng Yao; first paragraph drop cap 2 lines (0.2cm from text); remaining paragraphs (excluding table): left/right indent 1.5 characters, first-line indent 2 characters, 1-line spacing before; split third paragraph into 2 equal columns (width 15 characters, separator line); set fourth paragraph to ``Keep lines together.''

4. Add table caption ``公众号关注量统计表'' in size 18pt HuaWen CaiYun, bold, centered, dark red (standard); insert blank column on right labeled ``合计,'' fill with row-sum formulas; sort by ``合计'' column descending (numeric); center table, center first row and column content vertically and horizontally, right-align remaining cells; row height 0.6cm, first column width 2.5cm, other columns 1.8cm.

5. Shade first row ``Turquoise, Accent 5, Lighter 80\%''; set outer borders and first-row bottom border to red 0.75pt double line; other inner borders to red 0.5pt single line.
\end{quote}

\normalsize
\noindent\textbf{Scoring:} 71 criteria spanning text formatting and find/replace (35), table structure and formatting (22), page setup (10), and document properties (4). Each criterion checks a specific document property (e.g., ``title font is SimHei,'' ``page size is 18.4cm $\times$ 26cm,'' ``table row height is 0.6cm'').

\subsection{Excel task: Three-Year Temperature Statistics (20 criteria)}

\small
\begin{quote}
打开考生文件夹下的电子表格Excel.xlsx工作簿文件，按照下列要求完成对此表格的操作并保存。

1.选择Sheet1工作表，将A1:G1单元格合并为一个单元格，文字居中对齐；利用AVERAGE函数计算近三年每月平均高温的平均值置于"平均高温平均值"列（H4:H15，保留小数点后1位），计算近三年每月平均低温的平均值置于"平均低温平均值"列（I4:I15，保留小数点后1位）；利用MAX函数计算近三年每月平均高温的最高值置于"平均高温最高值"列（J4:J15），利用MIN函数计算近三年每月平均低温的最低值置于"平均低温最低值"列（K4:K15）。利用条件格式修饰B4:G15单元格区域，值大于28的单元格设置为"浅红填充色深红色文本"，值小于12的单元格设置为"绿填充色深绿色文本"；利用条件格式修饰H4:K15单元格区域，基于各自值设置所有单元格的格式为实心填充数据条，自定义颜色（RGB：216,194,246），条形图方向为从左到右。

2.选取Sheet1工作表中的"月份"列（A3:A15）、"平均高温平均值"列（H3:H15）、"平均低温平均值"列（I3:I15）、"平均高温最高值"列（J3:J15）和"平均低温最低值"列（K3:K15）数据区域的内容建立"折线图"，图表样式为"样式3"，图表布局为"布局5"，垂直坐标轴标题设置为"单位（度）"，图表标题为"近三年气温统计图"，将图表插入到当前工作表的"A17:H33"单元格区域内，将Sheet1工作表命名为"近三年气温统计表"。

3.选择"产品销售情况表"工作表，对工作表内数据清单的内容按主要关键字"分公司"的升序和次要关键字"季度"的升序进行排序；对排序后的数据进行筛选，条件：产品名称为电视机、电冰箱、数码相机和空调，且销售额排名小于或等于30，工作表名不变，保存Excel.xlsx工作簿。
\end{quote}

\normalsize
\noindent\textbf{English translation:}
\small
\begin{quote}
Open Excel.xlsx and complete the following:

1. On Sheet1, merge A1:G1 and center; use AVERAGE to compute three-year mean high temperatures (H4:H15, 1 decimal) and mean low temperatures (I4:I15, 1 decimal); use MAX for highest high temperatures (J4:J15) and MIN for lowest low temperatures (K4:K15). Apply conditional formatting to B4:G15: values $>$28 $\to$ ``Light Red Fill with Dark Red Text,'' values $<$12 $\to$ ``Green Fill with Dark Green Text.'' Apply data-bar conditional formatting to H4:K15: solid fill, custom color (RGB: 216, 194, 246), left-to-right direction.

2. Create a line chart from columns A3:A15, H3:H15, I3:I15, J3:J15, K3:K15; chart style ``Style 3,'' layout ``Layout 5''; vertical axis title ``单位（度）'' (``Unit (\textdegree C)''); chart title ``近三年气温统计图'' (``Three-Year Temperature Statistics''); place chart at A17:H33; rename sheet to ``近三年气温统计表.''

3. On the ``产品销售情况表'' (Product Sales) sheet, sort by ``分公司'' (Branch) ascending then ``季度'' (Quarter) ascending; filter: product name = TV, refrigerator, digital camera, or air conditioner, and sales rank $\leq$ 30.
\end{quote}

\normalsize
\noindent\textbf{Scoring:} 20 criteria covering formulas and cell content (5), conditional formatting (4), chart properties (6), worksheet tab name (1), and data sorting/filtering (4). Representative criteria include: ``AVERAGE formula is correct in H4,'' ``chart type is line chart,'' ``sheet tab name is 近三年气温统计表.''

\subsection{PowerPoint task: High Temperature Warning (21 criteria)}

\small
\begin{quote}
打开考生文件夹下的演示文稿yswg.pptx，按照下列要求完成对此文稿的修饰并保存。

1.为整个演示文稿应用"离子"主题，设置幻灯片的大小为"全屏显示（16:9）"，并在幻灯片大小变更时，按比例调整内容以适应新的幻灯片尺寸；放映方式为"观众自行浏览"。

2.在第一张幻灯片前插入一张版式为"标题幻灯片"的新幻灯片，主标题为"北京河北山东陕西等地7月6日高气温将达40℃"，副标题为"高温预警"。

3.第二张幻灯片版式改为"两栏内容"；标题为"高温黄色预警"；将考生文件夹下图片文件PPT1.PNG移到右侧内容区；左侧文本设置为"黑体"、23磅字；图片动画设置为"强调/陀螺旋"，效果选项为"数量/半旋转"。

4.在幻灯片的最后插入一张版式为"空白"的幻灯片，插入一个SmartArt图形，版式为"水平层次结构"，SmartArt样式为"砖块场景"，SmartArt图形中的所有文字从考生文件夹下的文本文件PPT2.txt中获取。

5.在幻灯片的最后插入一张版式为"标题和内容"的新幻灯片，标题为"高温防御指南"；内容区插入5行2列的表格，表格样式为"中度样式2"。第1行的1、2列内容依次为"有关单位和人员"和"高温防御措施"，其他单元格的内容根据考生文件夹下文本文件PPT3.txt内容按顺序依次从上到下填写，例如第2行的1、2列内容依次为"媒体"和"应加强防暑降温保健知识的宣传；"。表格内文字均设置为22磅字，并在备注区插入文本"全社会动员起来防御高温"。

6.全体幻灯片切换方式为"百叶窗"，效果选项为"水平"。
\end{quote}

\normalsize
\noindent\textbf{English translation:}
\small
\begin{quote}
Open yswg.pptx and complete the following:

1. Apply the ``Ion'' theme; set slide size to widescreen (16:9), scaling content to fit; set show type to ``Browsed by an Individual.''

2. Insert a new ``Title Slide'' before slide 1; main title ``北京河北山东陕西等地7月6日高气温将达40℃'' (``Beijing, Hebei, Shandong, Shaanxi forecast 40\textdegree C on July 6''); subtitle ``高温预警'' (``High Temperature Warning'').

3. Change slide 2 to ``Two Content'' layout; title ``高温黄色预警'' (``Yellow Heat Warning''); move PPT1.PNG to right content area; set left text to SimHei 23pt; set image animation to ``Emphasis/Spin,'' effect option ``Amount/Half Spin.''

4. Insert a blank slide at the end; add SmartArt ``Horizontal Hierarchy'' with ``Brick Scene'' style; populate text from PPT2.txt.

5. Insert a ``Title and Content'' slide at the end; title ``高温防御指南'' (``Heat Defense Guide''); insert a 5$\times$2 table with ``Medium Style 2''; fill header row with ``有关单位和人员'' and ``高温防御措施,'' remaining rows from PPT3.txt; set all table text to 22pt; add note ``全社会动员起来防御高温.''

6. Apply ``Blinds'' transition to all slides with ``Horizontal'' effect option.
\end{quote}

\normalsize
\noindent\textbf{Scoring:} 21 criteria covering transitions and animations (8), slide layout and content (7), tables (4), graphics/media (1), and document properties (1). These criteria verify properties such as ``theme name is Ion,'' ``show type is Browsed by an Individual,'' ``SmartArt layout is Horizontal Hierarchy,'' and ``transition type is Blinds/Horizontal.''
\end{CJK*}

\section{Detailed scoring configuration}
\label{app:scoring}

Table~\ref{tab:weptype_full} provides the complete mapping from criterion types (WepTypes) to our 8 skill categories.

\begin{table}[t]
  \caption{Complete mapping from evaluation criterion types (WepTypes) to skill categories. Count indicates the number of ScorePointDefine elements across all 200 tasks.}
  \label{tab:weptype_full}
  \centering
  \small
  \begin{tabular}{llr}
    \toprule
    \textbf{Skill Category} & \textbf{WepType(s)} & \textbf{Count} \\
    \midrule
    \multirow{5}{*}{Text \& Format} & ParaFind & 2,160 \\
    & ParaFont & 109 \\
    & TextEffect & 136 \\
    & FormatParaGraph & 51 \\
    & Body & 288 \\
    \midrule
    \multirow{4}{*}{Layout \& Design} & PageSetup & 483 \\
    & SlideMasterEffect & 81 \\
    & SlideFind & 648 \\
    & SlideCustom & 41 \\
    \midrule
    \multirow{2}{*}{Tables} & Table & 185 \\
    & TableTestRowCol & 586 \\
    \midrule
    Charts & Chart & 362 \\
    \midrule
    \multirow{4}{*}{Data \& Formulas} & PivotTable & 91 \\
    & FormatCondition & 63 \\
    & Subtotal & 14 \\
    & Border & 3 \\
    \midrule
    \multirow{5}{*}{Graphics \& Media} & ShapePicExist & 237 \\
    & SmartArt & 293 \\
    & FormatPicture & 58 \\
    & Shape & 31 \\
    & TextFrameFind & 30 \\
    \midrule
    \multirow{2}{*}{Animation} & AnimationSetting & 178 \\
    & Transition & 377 \\
    \midrule
    \multirow{6}{*}{Doc.\ Properties} & Document & 174 \\
    & Common & 220 \\
    & WorksheetTab / WorkSheetTab & 73 \\
    & Sheet & 18 \\
    & Workbook & 8 \\
    & Custom / CustShow / ParseAllProps & 120 \\
    \bottomrule
  \end{tabular}
\end{table}

\section{Additional results}
\label{app:additional}

\subsection{Per-task score distribution}

Table~\ref{tab:score_dist} summarizes the per-task score distribution for each model, computed from three-run averages.
Scores are highly skewed: most models' first quartile is 0\%, reflecting the large fraction of tasks that receive zero credit (typically due to code execution failure).
Claude Opus 4.7 achieves a median of 32.1\%, meaning half of all tasks score below this mark, and is one of only two models to achieve a near-perfect score on any task.
Weaker models are dominated by zero-score outcomes: Qwen3.5-397B-A17B receives exactly 0 on more than half of all tasks (104 out of 200), resulting in a median of 0\%.
Standard deviations are large (6.8--25.7 percentage points), indicating high variance across tasks: some tasks are solvable by every model while others defeat all of them.

\begin{table}[t]
  \caption{Per-task score distribution (\%) across models, computed from three-run averages. Q25/Q50/Q75 denote quartiles. ``Zeros'' counts tasks receiving exactly 0/100 averaged across runs.}
  \label{tab:score_dist}
  \centering
  \small
  \begin{tabular}{l rrrrrrr}
    \toprule
    \textbf{Model} & \textbf{Mean} & \textbf{Std} & \textbf{Q25} & \textbf{Q50} & \textbf{Q75} & \textbf{Max} & \textbf{Zeros} \\
    \midrule
    GPT-5.5 & 36.1 & 25.6 & 16.7 & 31.5 & 58.0 & 90.8 & 21 \\
    Claude Opus 4.7 & 36.6 & 25.7 & 15.9 & 32.1 & 55.8 & 96.0 & 20 \\
    Gemini 3.1 Pro & 16.3 & 18.9 & 0.0 & 9.3 & 26.7 & 80.7 & 61 \\
    Kimi-K2.6 & 9.4 & 15.3 & 0.0 & 1.5 & 13.4 & 74.3 & 90 \\
    Qwen3.5-397B-A17B & 3.9 & 7.0 & 0.0 & 0.0 & 5.0 & 40.8 & 104 \\
    MiMo-V2.5 & 3.5 & 7.1 & 0.0 & 0.0 & 4.2 & 50.0 & 108 \\
    Grok-4.1-fast & 2.8 & 6.8 & 0.0 & 0.0 & 2.5 & 50.0 & 127 \\
    \bottomrule
  \end{tabular}
\end{table}

\subsection{Complete per-skill results}

Table~\ref{tab:perskill_full} presents per-skill criterion pass rates for all seven evaluated models.
Tables, Charts, and Data \& Formulas show the strongest peak rates for top models (up to 59.1\%), indicating that table structure, formula-based operations, and chart operations are comparatively amenable to code generation.
Animation and Graphics \& Media are consistently the weakest categories, reflecting the limited API coverage of \texttt{python-pptx} for transitions, effects, and SmartArt.
The four weakest models (Kimi-K2.6 through Grok-4.1-fast) show relatively uniform profiles with rates below 20\% across all categories, while stronger models exhibit more pronounced peaks and valleys.
GPT-5.5 leads in criterion pass rate on Charts, Data \& Formulas, Graphics \& Media, and Animation, while Claude Opus 4.7 leads on Text \& Format, Layout \& Design, Tables, and Doc.\ Properties. Despite comparable task-level SR (36.2\% vs.\ 36.6\%) and overall criterion pass rates (40.3\% vs.\ 40.5\%), the two models exhibit complementary strengths: GPT-5.5 excels at data-centric and chart operations, while Claude Opus 4.7 is stronger at document structure and table formatting.

\begin{table}[t]
  \caption{Per-skill criterion pass rate (\%) for all seven evaluated models. Bold indicates the highest rate per skill. Criteria are classified into 8 skill categories based on their evaluation type (WepType); see Table~\ref{tab:weptype_full} for the full mapping.}
  \label{tab:perskill_full}
  \centering
  \small
  \setlength{\tabcolsep}{3pt}
  \resizebox{\textwidth}{!}{%
  \begin{tabular}{l rrr rrrr}
    \toprule
    \textbf{Skill Category} & \textbf{Claude Opus 4.7} & \textbf{GPT-5.5} & \textbf{Gemini 3.1 Pro} & \textbf{Kimi-K2.6} & \textbf{Qwen3.5-397B-A17B} & \textbf{MiMo-V2.5} & \textbf{Grok-4.1-fast} \\
    \midrule
    Text \& Format (2744) & \textbf{46.6} & 45.8 & 16.8 & 15.1 & 4.8 & 4.5 & 4.1 \\
    Layout \& Design (1253) & \textbf{39.8} & 37.4 & 9.4 & 9.7 & 6.1 & 4.8 & 6.1 \\
    Tables (771) & \textbf{59.1} & 44.0 & 9.7 & 10.8 & 6.1 & 5.8 & 4.2 \\
    Charts (362) & 36.7 & \textbf{51.4} & 34.0 & 19.1 & 6.1 & 6.4 & 5.5 \\
    Data \& Formulas (171) & 43.9 & \textbf{48.5} & 39.8 & 18.7 & 6.4 & 8.8 & 7.0 \\
    Graphics \& Media (649) & 25.1 & \textbf{29.1} & 10.2 & 7.9 & 7.4 & 7.7 & 7.9 \\
    Animation (555) & 21.4 & \textbf{26.7} & 4.0 & 4.0 & 2.3 & 3.8 & 3.6 \\
    Doc.\ Properties (613) & \textbf{51.1} & 39.5 & 22.8 & 13.4 & 11.1 & 8.8 & 9.6 \\
    \midrule
    Overall (7118) & \textbf{40.5} & 40.3 & 18.3 & 12.3 & 6.3 & 6.3 & 6.0 \\
    \bottomrule
  \end{tabular}%
  }
\end{table}

\subsection{Task difficulty analysis}

To assess the discriminative power of individual tasks, we analyze the score distribution across all 7 models.
Of 200 tasks, 1 (0.5\%) defeats all models (every model's three-run average is 0\%), while 67 (33.5\%) are partially solved by all models (every model scores above 0\%).
The remaining 132 tasks discriminate between model tiers: 34 tasks are solved only by the top models (GPT-5.5 or Claude Opus 4.7 scores $>$50\%) while the three weakest models all score 0\%.
The most discriminating tasks exhibit spreads of up to 100 percentage points between the best and worst models, confirming that \name provides meaningful separation across the full capability spectrum.
These statistics indicate a well-calibrated difficulty distribution: the benchmark is neither trivially easy nor impossibly hard, with the majority of tasks providing useful signal for distinguishing model capabilities.

\section{Reproducibility details}
\label{app:reproducibility}

\subsection{Agent prompts}

The single-turn baseline uses the following system prompt for all models:

\begin{quote}
\small
\texttt{You are an expert in Microsoft Office automation. Your task is to write Python code to manipulate Office documents according to the user's requirements.}

\texttt{Rules:}
\texttt{1. Write complete, directly executable Python code based on the task requirements.}
\texttt{2. Use python-docx for Word documents (.docx), openpyxl for Excel (.xlsx), and python-pptx for PowerPoint (.pptx).}
\texttt{3. The document content and task instructions are in Chinese.}
\texttt{4. Output ONLY the Python code, wrapped in \textasciigrave\textasciigrave\textasciigrave python and \textasciigrave\textasciigrave\textasciigrave. No explanations.}
\end{quote}

The user message includes: (1) task instructions from the examination (Chinese text with any embedded images), (2) screenshots of input document pages/sheets/slides rendered as JPEG images (max 1000px), and (3) a file listing with code generation instructions.
No few-shot examples are provided.

For the coding agents (Claude Code and Codex), the following prompt is provided directly without a custom system prompt:

\begin{quote}
\small
\texttt{Read the .rtf file in this directory for the task requirements. The input files to modify are in ./input/. Complete all the operations described and save all changes back to the original files.}
\end{quote}

The agent operates in an isolated temporary directory containing only the raw exam paper (RTF file) and the input document(s).
No pre-processed task text, document screenshots, library recommendations, or scoring rubrics are provided.
Claude Code runs with the \texttt{claude} CLI in non-interactive mode; Codex runs with the \texttt{codex} CLI. Both use no turn limit and a one-hour timeout per task.

\subsection{Decoding settings}

All models are called with \texttt{temperature=1.0}. GPT-5.5 uses an internal \texttt{temperature=1.0} setting that is not user-configurable, so we set all other configurable models to the same value for consistency.
To account for the resulting stochasticity, all main-paper results are means of 3 independent runs per model.

\subsection{Evaluation environment}

The evaluation engine primarily parses ISO/IEC 29500 Office Open XML structures using the Open XML SDK~\citep{iso29500,microsoft2024openxmlsdk}, with a subset of checks invoking Office COM automation, and evaluates each scoring criterion defined in the task's XML configuration.
For each criterion, the engine locates the specified document property, compares it against the expected value using the declared operator, and emits a binary pass/fail result.
The per-task score (0--100) is the sum of points from passing criteria.
The engine is implemented as a command-line tool; a 60-second timeout is applied per task.

\subsection{Library versions}

\begin{table}[t]
  \caption{Library versions used in experiments.}
  \label{tab:library_versions}
  \centering
  \small
  \begin{tabular}{ll}
    \toprule
    \textbf{Component} & \textbf{Version} \\
    \midrule
    Python & 3.12.10 \\
    python-docx & 1.2.0 \\
    openpyxl & 3.1.5 \\
    python-pptx & 1.0.2 \\
    Open XML SDK & 2.5 \\
    \bottomrule
  \end{tabular}
\end{table}

\subsection{Community solution error analysis}
\label{sec:community_errors}

The community reference solutions achieve 95.5\% on average, leaving a small gap from perfect scores.
We manually inspected the lowest-scoring community reference solutions and found that the deductions correspond to genuinely missing or incorrect operations (e.g., omitted page setup steps, incorrect font sizes, or missing chart elements).
In all inspected cases, the scoring engine correctly identified the discrepancy between the submitted document and the rubric specification.
This supports the use of the community reference solutions as a sanity check for task solvability and scoring behavior, while not making them a formal human baseline.



\end{document}